%% file: main.tex
\documentclass[10pt,twocolumn,letterpaper]{article}
\usepackage[T1]{fontenc}
\usepackage[pagenumbers]{cvpr}
\input{preamble}
\definecolor{linkblue}{rgb}{0.21,0.49,0.74}
\usepackage[pagebackref,breaklinks,colorlinks,allcolors=linkblue]{hyperref}
\title{From Matching to Generation: Learning Hierarchical Semantic Motion IDs for Generative Text-to-Motion Retrieval}
\author{%
  Yizhou Chen\textsuperscript{1},
  Guangtao Lyu\textsuperscript{2},
  Chenghao Xu\textsuperscript{3},
  Jiexi Yan\textsuperscript{1},
  Yanhua Yang\textsuperscript{1}\\[0.35em]
  \normalsize \textsuperscript{1}School of Computer Science and Technology, Xidian University, Xi'an, China\\
  \normalsize \textsuperscript{2}School of Electronic Engineering, Xidian University, Xi'an, China\\
  \normalsize \textsuperscript{3}College of Information Science and Engineering, Hohai University, Changzhou, China
}
\begin{document}
\maketitle
\input{sec/0_abstract}
\input{sec/1_intro}
\input{sec/2_related_work}
\input{sec/3_method}
\input{sec/4_experiments}
\input{sec/5_conclusion}

{\small
\bibliographystyle{ieeenat_fullname}
\bibliography{references}
}
\end{document}

%% file: preamble.tex
\usepackage{placeins}

%% file: sec/0_abstract.tex
\begin{abstract}
Text-to-motion retrieval aims to identify human motion sequences that match a natural language description. Existing methods primarily rely on cross-modal similarity matching in a shared embedding space, which limits deep cross-modal interaction during matching and hinders precise alignment of semantic details across modalities. Generative retrieval offers an alternative way to model relevance by enabling dynamic interaction between queries and candidate identifiers during generation, progressively refining candidate selection. However, adapting this paradigm to human motion poses two challenges: learning hierarchical semantic IDs that organize related motions while preserving fine-grained differences, and aligning ID generation with retrieval ranking. To address these challenges, we propose HiMoID, a generative framework for text-to-motion retrieval. The first stage, hierarchical semantic motion quantization, combines full-motion reconstruction with semantic prefix organization to preserve motion details and encourage semantically related motions to share prefixes. Building on this hierarchy, the second stage, retrieval-oriented semantic ID generation, uses query-conditioned branch targets and tree-structured distribution supervision to guide the generator's allocation of probability across candidate branches according to their relevance to the query. Extensive experiments on HumanML3D and KIT-ML demonstrate that HiMoID achieves state-of-the-art performance in text-to-motion retrieval.
\end{abstract}

%% file: sec/1_intro.tex
\section{Introduction}
\label{sec:intro}

Understanding human motion is a fundamental problem in computer vision and machine intelligence~\cite{zhu2023motionbert}. Natural language provides an intuitive interface for describing complex movements and searching motion collections. Given a textual description, text-to-motion retrieval aims to identify matching motion sequences from a gallery~\cite{messina2023mot,petrovich2023tmr}. The task supports applications in motion understanding~\cite{li2025lamp,lyu2025lex}, controllable motion generation~\cite{zhang2023remodiffuse,yu2025remogpt}, and motion editing~\cite{zhang2025kinmo}. Across these applications, effective retrieval depends on matching the movements described in the query.

\begin{figure*}[t]
  \centering
  \includegraphics[width=\textwidth]{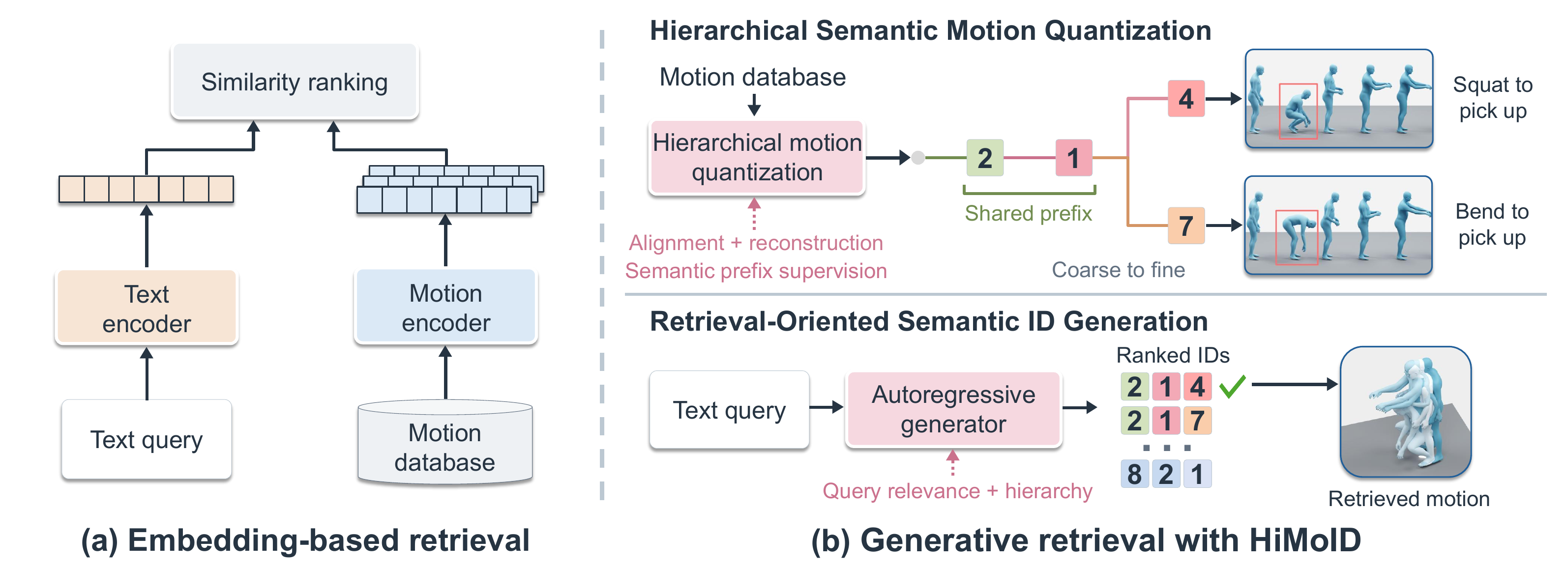}
  \caption{Comparison of text-to-motion retrieval paradigms. (a) Embedding-based retrieval ranks motions by text--motion similarity in a shared embedding space. (b) Our generative retrieval combines hierarchical semantic IDs with retrieval-oriented generation supervision.}
  \label{fig:motivation}
\end{figure*}

As illustrated in Fig.~\ref{fig:motivation}(a), conventional embedding-based methods encode the text query and each candidate motion independently as fixed-dimensional vectors, then rank the motions by cosine similarity. With independent encoding, much of the interaction between the query and a candidate occurs through a single similarity score~\cite{ziems2023llmurl}. This limits how directly the model can match individual details in the query to their counterparts in the motion. Generative retrieval offers an alternative by formulating candidate selection as query-conditioned identifier generation~\cite{tay2022dsi}. During generation, cross-attention allows the decoder to attend to different parts of the query based on the current prefix~\cite{wu2024grmvdr}. With hierarchical semantic identifiers~\cite{rajput2023tiger}, this process can first locate groups of motions that share an overall action and then distinguish variations in its execution.

Adapting quantization-based identifiers~\cite{rajput2023tiger,zhang2024irgen} and standard autoregressive training to text-to-motion retrieval presents two challenges. The first concerns what the identifiers encode. Motions with similar overall semantics may differ in local pose or temporal dynamics, and directly quantizing globally aligned representations can obscure these differences. Moreover, selecting codewords based on distances in the representation space leaves the semantic organization of shared prefixes implicit. Coarse-to-fine retrieval benefits from longer shared prefixes among more closely related motions, so that the generator can narrow the search before resolving finer differences. The second challenge concerns how the generator ranks candidates. The standard autoregressive objective increases the likelihood of the annotated identifier, while retrieval requires ranking candidate motions by their relevance to the query~\cite{li2024dgr}. Training on the annotated path alone leaves the relative relevance of other candidates unspecified, including how that relevance should shape the probability assigned to their branches.

To address these challenges, we propose HiMoID, a generative framework for text-to-motion retrieval, as illustrated in Fig.~\ref{fig:motivation}(b). The first stage, hierarchical semantic motion quantization, combines text--motion alignment, full-motion reconstruction, and semantic prefix organization to learn motion IDs that capture both shared action semantics and fine-grained differences. In the example in Fig.~\ref{fig:motivation}(b), two motions that involve picking up and placing an object share a prefix that captures their common action, while their final tokens distinguish squatting from bending at the waist. Building on this hierarchy, the second stage, retrieval-oriented semantic ID generation, uses query-conditioned branch targets and tree-structured distribution supervision to guide the generator in assigning probability to candidate branches according to their relevance to the query. At inference, trie-constrained decoding generates valid semantic IDs, which are mapped to motions through an offline index. By combining hierarchical motion quantization with retrieval-oriented generation, HiMoID enables precise text-to-motion retrieval.

In summary, our contributions are as follows:
\begin{itemize}
  \item We formulate text-to-motion retrieval as query-conditioned generation of hierarchical semantic IDs. This requires learning IDs that organize semantically related motions while preserving fine-grained differences, and aligning ID generation with retrieval ranking.
  \item We propose HiMoID, a generative framework that couples hierarchical semantic motion quantization with retrieval-oriented semantic ID generation. The quantization stage organizes motions into a coarse-to-fine hierarchy, which the generation stage uses to rank candidates by their relevance to the query.
  \item We conduct extensive experiments on HumanML3D and KIT-ML, demonstrating that HiMoID consistently outperforms existing text-to-motion retrieval methods.
\end{itemize}

%% file: sec/2_related_work.tex
\section{Related Work}
\label{sec:related_work}

\begin{figure*}[t]
  \centering
  \includegraphics[width=0.95\textwidth]{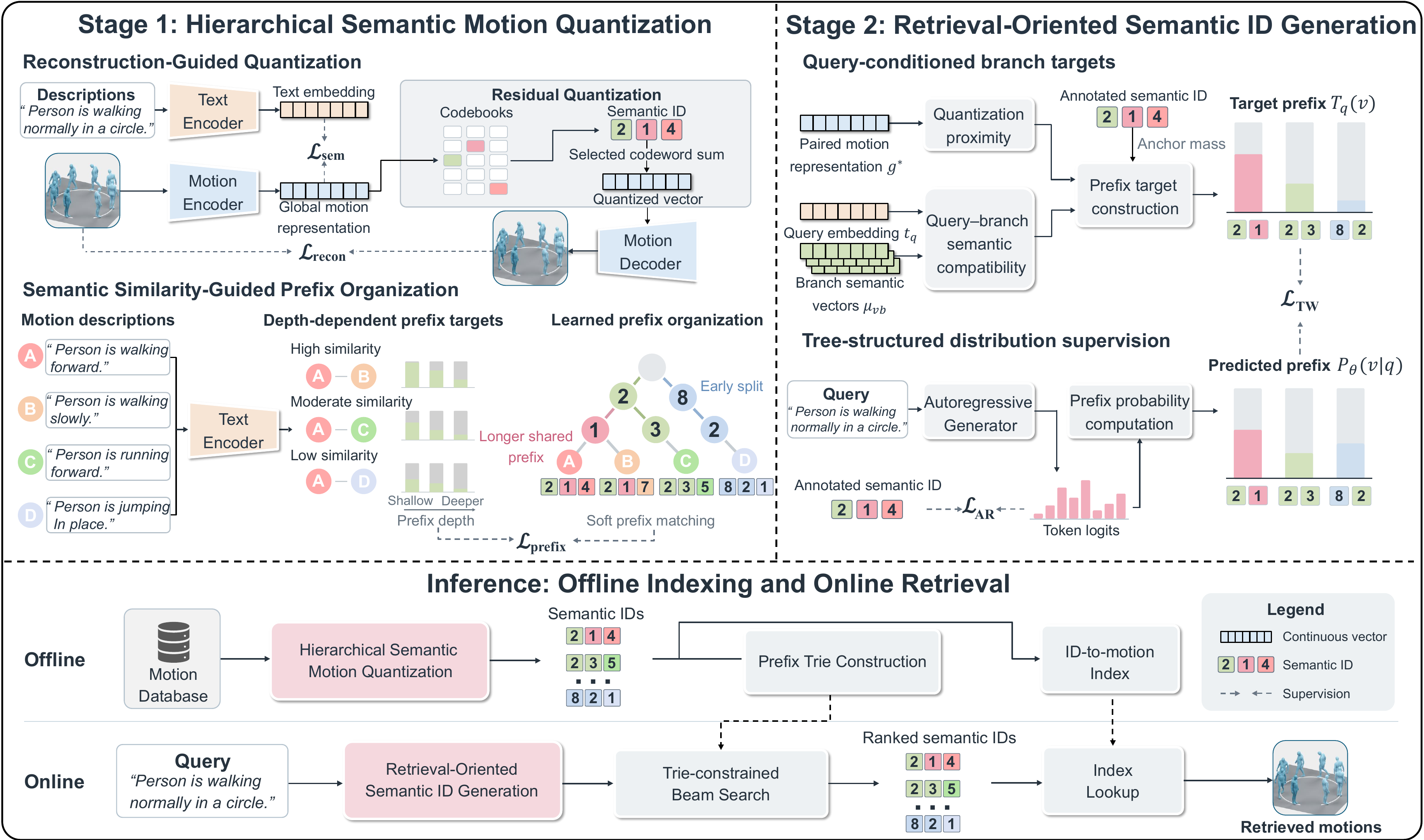}
  \caption{Overview of the proposed HiMoID. Stage 1 learns semantic IDs through reconstruction-guided motion quantization and semantic similarity-guided prefix organization. Stage 2 combines quantization proximity with query-branch semantic compatibility to construct target probabilities and trains an autoregressive generator with cross-entropy and tree-Wasserstein supervision. At inference, gallery motions are indexed offline, and queries retrieve motions via trie-constrained semantic ID generation and index lookup.}
  \label{fig:mr:framework}
\end{figure*}

\subsection{Text-to-Motion Retrieval}

Existing text-to-motion retrieval methods typically employ contrastive learning to align text and motion in a shared embedding space~\cite{messina2023mot,petrovich2023tmr}. MotionPatches and VLMA leverage pretrained vision--language models or video as an intermediate modality to strengthen cross-modal representations~\cite{yu2024motionpatches,yin2024vlma}. Sparse lexical representations further map both modalities into a shared lexical space, assigning explicit semantic meanings to representation dimensions and improving interpretability~\cite{lyu2025lex}. Fine-grained methods further improve local motion--text correspondences through hierarchical semantic interactions~\cite{yang2024hsa,yang2024mgsi,shi2024mesm} and modeling of body parts, temporal order, and kinematic structure~\cite{fujiwara2024car,zhang2025kinmo,zhang2025sgar,chen2026pst,ren2026wamo}. Despite their varied representations and alignment strategies, these approaches still formulate text-to-motion retrieval as candidate matching and ranking. We explore a generative formulation of this task, autoregressively generating motion semantic identifiers conditioned on the query to retrieve relevant sequences from the gallery.

\subsection{Generative Retrieval}
Generative retrieval predicts identifiers of relevant candidates directly from queries. GENRE generates entity names~\cite{decao2021genre}, and DSI learns indexing and retrieval through document ID prediction~\cite{tay2022dsi}. GenRet learns semantic IDs through discrete auto-encoding~\cite{sun2023genret}; TIGER adopts residual-quantized IDs for recommendation~\cite{rajput2023tiger}. IRGen and GENIUS extend identifier generation to image retrieval and multimodal search, respectively~\cite{zhang2024irgen,kim2025genius}. However, directly adapting these methods to motion presents challenges in preserving both movement details and semantic hierarchy, while standard generation training is not fully aligned with retrieval ranking. We therefore combine semantic alignment, motion reconstruction, and prefix supervision to learn motion identifiers, and guide generator training with query relevance and hierarchical candidate relationships.

%% file: sec/3_method.tex
\section{Method}
\label{sec:mr:method}

\subsection{Problem Formulation and Overview}
\label{sec:mr:overview}

Text-to-motion retrieval aims to retrieve semantically matching motion sequences from a motion gallery based on a natural-language description. Specifically, given a text query \(q\) and a motion gallery \(\mathcal G=\{X_i\}_{i=1}^{N}\), where \(X_i\in\mathbb R^{T_i\times d_x}\) and \(T_i\) and \(d_x\) denote the number of frames and the feature dimension per frame, respectively, the goal is to rank candidate motions by their relevance to the query. Under the generative retrieval paradigm \cite{rajput2023tiger,zhang2024irgen}, each motion is encoded as a semantic identifier (semantic ID) \(\mathbf c_i=(c_i^{(1)},\ldots,c_i^{(L)})\) of length \(L\), and the model selects candidates by generating semantic IDs conditioned on the query.

As illustrated in Fig.~\ref{fig:mr:framework}, we propose HiMoID, a two-stage generative framework comprising Hierarchical Semantic Motion Quantization and Retrieval-Oriented Semantic ID Generation. The first stage applies text-motion semantic alignment to global motion representations before quantization and reconstructs full motions from the quantized representations, encouraging semantic IDs to retain both global semantics and motion details for fine-grained retrieval. It also estimates semantic similarity from motion descriptions and supervises prefix sharing at different depths to organize candidates into coarse-to-fine groups. The second stage fixes the learned semantic IDs, combines the annotated motion's quantization proximity with query semantics to construct branch targets, and uses the tree-Wasserstein loss to supervise the generator's probability allocation on the prefix tree, aiming to better align generation supervision with query relevance and hierarchical relationships among candidates. At inference, the generator produces valid gallery semantic IDs under prefix-tree constraints and retrieves the corresponding motions through an offline index.

\subsection{Hierarchical Semantic Motion Quantization}
\label{sec:mr:quantization}

To support coarse-to-fine retrieval, we introduce hierarchical semantic motion quantization, which preserves fine-grained motion details in the learned semantic IDs and groups semantically related motions by shared prefixes.

\noindent\textbf{Reconstruction-Guided Quantization.}\  Global text-motion alignment provides motion representations with query-related semantics. However, quantizing only these coarse global representations learned through alignment may overlook local poses and temporal variations, making motions with similar overall semantics difficult to distinguish. We therefore combine pre-quantization semantic alignment with post-quantization full-motion reconstruction, enabling semantic IDs to retain both query-related semantics and fine-grained motion differences.

Given motion \(X_i\), the motion encoder produces global representation \(g_i=E_m(X_i)\in\mathbb R^{d_g}\). Motion and text are projected into normalized shared-space embeddings \(f_i\) and \(t_q\), aligned using a bidirectional multi-positive contrastive loss \(\mathcal L_{\mathrm{sem}}\) \cite{petrovich2023tmr,miech2020milnce}. Following generative retrieval methods \cite{rajput2023tiger,zhang2024irgen}, residual quantization \cite{lee2022rqvae} encodes \(g_i\) as semantic ID \(\mathbf c_i\) and quantized representation \(z_i\), using standard commitment regularization, EMA updates, and straight-through estimation \cite{kim2025genius,oord2017vqvae}.

To incorporate local motion content into ID learning, the motion decoder reconstructs the full sequence from the quantized representation and valid-length information. We constrain both overall motion features and local joint configurations through

\begin{equation}
\label{eq:mr:reconstruction}
\mathcal L_{\mathrm{recon}}
=\lambda_{\mathrm{rec}}\ell_{\mathrm{rec}}
+\lambda_{\mathrm{pos}}\ell_{\mathrm{pos}},
\end{equation}
where \(\ell_{\mathrm{rec}}\) and \(\ell_{\mathrm{pos}}\) denote Smooth L1 errors \cite{girshick2015fast} on full motion features and local joint positions, respectively. Full-sequence reconstruction requires quantized representations to retain temporally varying motion content, while the position term further emphasizes pose differences, providing supervision for distinguishing semantically similar candidates. Gradients reach the motion encoder through the straight-through estimator, and the reconstruction branch is used only during quantization training.

\noindent\textbf{Semantic Similarity-Guided Prefix Organization.}\  Full-motion reconstruction constrains ID content but does not determine how long a prefix different motions should share. To let the generator first locate candidate groups by common semantics and then progressively distinguish finer movement differences, we construct depth-dependent prefix matching targets from motion descriptions.

Specifically, a text teacher \(\bar E_t\) aggregates each motion's descriptions into a semantic representation, and a Gaussian kernel \cite{ng2001spectral} computes motion-pair semantic similarity \(a_{ij}\in(0,1]\) from squared Euclidean distances between these representations. For the first \(L_s<L\) levels, let \(t_{ij}^{(m)}\) denote the matching target for the first \(m\) levels of motions \(i\) and \(j\), defined as

\begin{equation}
\label{eq:mr:depthtarget}
t_{ij}^{(m)}=a_{ij}^{\gamma_m},
\end{equation}
where the exponents satisfy \(0<\gamma_1\leq\cdots\leq\gamma_{L_s}\). Since similarity is at most 1, increasing the exponent makes the target for a given pair non-increasing. Moderately similar motions can therefore have relatively high shallow-prefix targets that decrease at deeper levels, while highly similar motions can retain high targets further down the hierarchy. This encourages semantically closer motions to share longer prefixes, leaving unsupervised levels free to express motion differences.

To use these targets for encoder training, we must address the lack of direct gradients through discrete prefix equality. Using motion \(j\)'s current ID as a reference, we match motion \(i\) by subtracting codewords already selected on the reference prefix, giving residual \(r_{i\mid j}^{(l)}=g_i-\sum_{h<l}e_{c_j^{(h)}}^{(h)}\). Let \(e_k^{(l)}\) denote a level-\(l\) codeword. Applying Softmax to negative distances \(-\|r_{i\mid j}^{(l)}-e_k^{(l)}\|_2^2/(d_g\tau_l^{\mathrm{match}})\) over all codewords at that level gives assignment probabilities \(\pi_{i\mid j}^{(l)}(k)\), with temperature \(\tau_l^{\mathrm{match}}>0\).

Multiplying the probabilities of the corresponding codewords along the reference prefix gives matching score \(p_{i\rightarrow j}^{(m)}=\prod_{l=1}^{m}\pi_{i\mid j}^{(l)}(c_j^{(l)})\). Swapping the two motions' roles and averaging yields a symmetric score:

\begin{equation}
\label{eq:mr:symmetric}
\widetilde p_{ij}^{(m)}
=\frac{p_{i\rightarrow j}^{(m)}+p_{j\rightarrow i}^{(m)}}{2}.
\end{equation}

Aligning this score with the semantic target translates description-derived relationships into training constraints on prefix organization:

\begin{equation}
\label{eq:mr:prefix}
\mathcal L_{\mathrm{prefix}}
=\mathbb E_{i,j,m}
\operatorname{BCE}\!\left(t_{ij}^{(m)},\widetilde p_{ij}^{(m)}\right),
\end{equation}
where the expectation uses balanced motion-pair weights and averages uniformly over the first \(L_s\) levels. Teacher targets, reference indices, and codeword vectors receive no gradients from this loss, so prefix supervision updates only the motion encoder through motion representations.

\subsection{Retrieval-Oriented Semantic ID Generation}
\label{sec:mr:generation}

Quantization provides motion IDs with a semantic hierarchy, whereas standard single-target autoregressive training increases the annotated path's probability without explicitly constraining how other candidates' relevance should affect branch probabilities. To narrow this gap between single-target generation supervision and retrieval objectives, we first construct a query-conditioned target distribution and then supervise the generator through a distributional distance on the ID tree.

\noindent\textbf{Query-conditioned branch targets.}\  To supervise potentially relevant candidates beyond the annotated path, targets should reflect both motion proximity to the annotated motion and compatibility with the current query. The former uses motion information from the annotation, while the latter makes branch selection reflect the semantics emphasized by the text.

We build prefix tree \(\mathcal T\) from training IDs. Let \(\operatorname{Ch}(v)\) denote valid children of prefix \(v\), and \(vb\) the child prefix formed by appending code index \(b\). Motions sharing a prefix form a candidate group; we average their semantic vectors \(f_i\) and normalize the result to obtain branch representation \(\mu_v\). The query representation remains \(t_q\). For training query \(q\), let \(X^*\) denote its paired annotated motion, with global representation \(g^*=E_m(X^*)\). Prefix \(v\) at depth \(l-1\) has residual \(r_v\), obtained by subtracting its selected codewords from \(g^*\). A valid child \(b\in\operatorname{Ch}(v)\) is scored as

\begin{equation}
\label{eq:mr:branch}
a_q(b\mid v)
=-\frac{d(r_v,e_b^{(l)})}{\tau_l^{\mathrm{branch}}}
+\eta_l\,t_q^\top\mu_{vb},
\end{equation}
where \(d(r,e)=\|r-e\|_2^2/d_g\) is the normalized squared distance, \(e_b^{(l)}\) is a level-\(l\) codeword, \(\tau_l^{\mathrm{branch}}>0\) is a temperature, and \(\eta_l\geq0\) weights query semantics. The first term measures quantization proximity to the annotated motion, while the second measures query-group semantic compatibility. Applying Softmax over all valid children gives conditional probabilities, whose product along each path yields target prefix probability \(R_q(v)\).

While extending supervision to other candidates, we retain explicit supervision for the annotated motion. We reserve a fraction \(\alpha\in[1/2,1)\) of the complete-ID distribution for the annotated ID \(\mathbf c^*\), allocating the remaining probability mass according to \(R_q\):

\begin{equation}
\label{eq:mr:target}
T_q(v)=\alpha\,\mathbf{1}[v\preceq\mathbf c^*]+(1-\alpha)R_q(v),
\end{equation}
where \(v\preceq\mathbf c^*\) indicates that \(v\) is a prefix of the annotated ID. This assigns the complete annotated ID probability of at least \(\alpha\) and ensures that each parent's probability equals the sum of its children's probabilities. Supervision at all depths therefore comes from one distribution, retaining the annotated target while expressing relevance preferences for other candidates. For brevity, \(a_q,R_q,T_q\) omit dependence on paired motion \(X^*\).

\noindent\textbf{Tree-structured distribution supervision.}\  Beyond constraining probability magnitudes, supervision should distinguish which candidate groups receive erroneous probability. Allocating probability to a motion sharing a long prefix should incur a different cost from allocating it to one that diverges early. We therefore use Wasserstein-1 distance under a tree metric \cite{le2019treesliced} to measure distributional errors according to hierarchical relationships among IDs.

Target \(T_q(v)\) is the total probability of all complete IDs under prefix \(v\), whereas each generator step predicts a conditional probability. To compare the same candidate group, we normalize generator predictions over valid children and multiply along paths to obtain prefix probability \(P_\theta(v\mid q)\). The tree-Wasserstein loss is

\begin{equation}
\label{eq:mr:tree}
\mathcal L_{\mathrm{TW}}(q,X^*)
=\sum_{v\ne\varnothing}
w_v\left|P_\theta(v\mid q)-T_q(v)\right|,
\end{equation}
where \(w_v>0\) denotes the fixed edge length linking prefix \(v\) to its parent, and the sum covers all non-root prefixes. Accumulating each subtree's total probability difference along its edge constrains both coarse candidate groups and finer branches. With equal edge lengths at each depth, earlier-diverging IDs have longer connecting paths and higher redistribution costs. To control computation, training retains the complete annotated path and preferentially expands other branches with higher target probabilities. Expanded nodes still normalize over all valid children, while subtrees that are not further expanded retain their total probability mass. Targets and predictions are compared on the same truncated tree, yielding a coarser distance after merging subtrees.

\begin{table*}[t]
  \centering
  \caption{Text-to-motion retrieval recall (\%) on HumanML3D and KIT-ML. Bold marks the best results and ``--'' denotes unreported values.}
  \label{tab:main-retrieval}
  \small
  \setlength{\tabcolsep}{4pt}
  \begin{tabular*}{\textwidth}{@{\extracolsep{\fill}}l*{10}{c}@{}}
    \toprule
    \textbf{Methods} & \multicolumn{5}{c}{HumanML3D} & \multicolumn{5}{c}{KIT-ML} \\
    \cmidrule(lr){2-6}\cmidrule(lr){7-11}
    & R@1$\uparrow$ & R@2$\uparrow$ & R@3$\uparrow$ & R@5$\uparrow$ & R@10$\uparrow$
    & R@1$\uparrow$ & R@2$\uparrow$ & R@3$\uparrow$ & R@5$\uparrow$ & R@10$\uparrow$ \\
    \midrule
    TEMOS~\cite{petrovich2022temos} & 2.12 & 4.09 & 5.87 & 8.26 & 13.52 & 7.11 & 13.25 & 17.59 & 24.10 & 35.66 \\
    T2M~\cite{guo2022humanml3d} & 1.80 & 3.42 & 4.79 & 7.12 & 12.47 & 3.37 & 6.99 & 10.84 & 16.87 & 27.71 \\
    TMR~\cite{petrovich2023tmr} & 8.92 & 12.04 & 16.33 & 22.06 & 33.37 & 10.05 & 13.87 & 20.74 & 30.03 & 44.66 \\
    MotionPatch~\cite{yu2024motionpatches} & 10.80 & 14.98 & 20.00 & 26.72 & 38.02 & 14.02 & 21.08 & 28.91 & 34.10 & 50.00 \\
    Lyu et al.~\cite{lyu2025lex} & 11.80 & 17.11 & 23.25 & 30.81 & 43.36 & 15.13 & 23.74 & 31.61 & 36.81 & 54.12 \\
    SGAR~\cite{zhang2025sgar} & 12.86 & -- & 23.52 & 30.75 & 43.00 & 16.75 & -- & 30.48 & 40.00 & 53.61 \\
    PST~\cite{chen2026pst} & 12.45 & 19.02 & 26.50 & 33.65 & 48.22 & 16.01 & 25.37 & 34.21 & 38.36 & 57.87 \\
    WaMo~\cite{ren2026wamo} & 14.02 & 17.58 & 25.51 & 32.06 & 42.10 & 18.31 & 24.82 & 34.46 & 43.49 & 56.75 \\
    HiMoID (ours) & \textbf{14.73} & \textbf{21.62} & \textbf{29.13} & \textbf{37.65} & \textbf{51.10}
         & \textbf{19.52} & \textbf{28.07} & \textbf{37.71} & \textbf{44.94} & \textbf{60.12} \\
    \bottomrule
  \end{tabular*}
\end{table*}

\subsection{Training and Inference}
\label{sec:mr:training}

\noindent\textbf{Quantization training.}\  We first perform cross-modal alignment and initialize codebooks from the aligned motion representations, giving the initial quantization partition a text-related semantic basis. To reduce interference from unstable early representations and codebooks, we initially train with semantic alignment, commitment, and reconstruction before progressively introducing prefix supervision. The text branch continues to learn through semantic alignment. Once prefix supervision is enabled, teacher targets are refreshed using the current text encoder and projection head at each epoch's start, then detached and fixed within that epoch.

Let \(\mathcal L_{\mathrm{base}}=\mathcal L_{\mathrm{sem}}+\lambda_{\mathrm{com}}\mathcal L_{\mathrm{com}}+\mathcal L_{\mathrm{recon}}\) denote the base objective, where \(\mathcal L_{\mathrm{com}}\) is the commitment loss. The full quantization objective is

\begin{equation}
\label{eq:mr:quantization}
\mathcal L_{\mathrm{ID}}
=\mathcal L_{\mathrm{base}}
+\lambda_{\mathrm{prefix}}\mathcal L_{\mathrm{prefix}}.
\end{equation}

\noindent\textbf{Generation training.}\  After learning IDs, we freeze the quantizer, cross-modal encoders, and projection heads, giving the generator fixed IDs and branch targets. Full-vocabulary autoregressive cross-entropy \(\mathcal L_{\mathrm{AR}}\) \cite{rajput2023tiger,kim2025genius} retains annotated-sequence supervision, while the tree-Wasserstein loss constrains hierarchical allocation among valid candidates. Together, they form

\begin{equation}
\label{eq:mr:generator}
\mathcal L_{\mathrm{gen}}
=\mathcal L_{\mathrm{AR}}
+\lambda_{\mathrm{TW}}\mathcal L_{\mathrm{TW}}(q,X^*).
\end{equation}

\noindent\textbf{Inference.}\  Gallery motions are encoded offline to build a prefix tree and motion index. Online, the generator uses trie-constrained beam search \cite{zhang2024irgen,decao2021genre} to restrict outputs to retrievable IDs, then ranks them by sequence scores and looks up corresponding motions. Reconstruction and target-construction teachers are used only during training, so online retrieval uses the generator and discrete index.

%% file: sec/4_experiments.tex
\section{Experiments}
\label{sec:experiments}

\subsection{Experimental Settings}
\label{sec:experimental-settings}

\noindent\textbf{Datasets.}\
We evaluate our method on two widely used public motion--language datasets: HumanML3D~\cite{guo2022humanml3d} and KIT-ML~\cite{plappert2016kitml}. Following TMR~\cite{petrovich2023tmr}, we use training, validation, and test splits with 23,384, 1,460, and 4,380 motions for HumanML3D and 4,888, 300, and 830 for KIT-ML.

\noindent\textbf{Evaluation Metrics.}\
To assess the retrieval performance of our method, we follow previous works \cite{petrovich2023tmr,yu2024motionpatches} and adopt recall at rank \(K\) (\(R@K\)) as our evaluation metric. \(R@K\) measures the percentage of queries with at least one correct motion among the top \(K\) results, where higher values indicate better performance. We report R@1, R@2, R@3, R@5, and R@10. We follow the multi-positive evaluation protocol of MotionPatch~\cite{yu2024motionpatches}. In addition, residual quantization does not guarantee distinct semantic IDs for different motions, so multiple motions may share the same semantic ID, resulting in a collision. To handle such collisions, generated semantic IDs are sorted by descending generation score and expanded into their corresponding gallery motions, with motions within each semantic ID bucket arranged in a fixed gallery-index order. Recall is computed on the resulting ranked motion list.

\noindent\textbf{Experimental Implementation.}\
We train our models on a single NVIDIA RTX A6000 GPU with 48~GB of memory. We use the MoMask convolutional backbone \cite{guo2024momask} as our motion encoder and pretrained DistilBERT \cite{sanh2019distilbert} as our text encoder. Global motion representations are discretized using a three-level residual quantizer with 256 codewords per level, yielding three-token semantic IDs. We fine-tune T5-base \cite{raffel2020t5} to generate these IDs from text queries. We set \(L_s=2\) and \(\alpha=0.75\). Training consists of 100 epochs of continuous cross-modal alignment, followed by 50 epochs of joint quantization and 30 epochs of generator training, with a batch size of 64 throughout. We use AdamW \cite{loshchilov2019adamw} for optimization. During joint quantization, the motion and text encoders use learning rates of \(10^{-5}\) and \(10^{-6}\), respectively; the generator is trained with a learning rate of \(10^{-4}\). The full-feature and local-position reconstruction weights are set to 0.1 and 0.05, respectively. The tree-Wasserstein loss weight is set to 0.1. At inference, we perform prefix-tree-constrained beam search with a beam size of 20.

\subsection{Experimental Results}
\label{sec:comparison}

\noindent\textbf{Performance Comparison.}\
To evaluate the text-to-motion retrieval performance of our framework, we conduct comparative experiments on HumanML3D and KIT-ML. We compare our method with previous works, including TEMOS~\cite{petrovich2022temos}, T2M~\cite{guo2022humanml3d}, TMR~\cite{petrovich2023tmr}, MotionPatch~\cite{yu2024motionpatches}, Lyu et al.~\cite{lyu2025lex}, SGAR~\cite{zhang2025sgar}, PST~\cite{chen2026pst}, and WaMo~\cite{ren2026wamo}. As shown in Table~\ref{tab:main-retrieval}, HiMoID outperforms all prior methods across all reported recall metrics on both datasets, demonstrating the effectiveness of our generative retrieval framework.

\begin{figure}[!t]
  \centering
  \includegraphics[width=\linewidth]{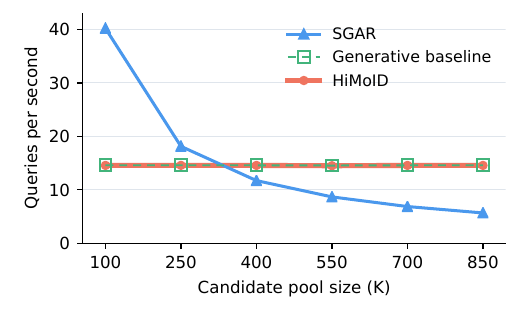}
  \caption{Online retrieval throughput on HumanML3D as the candidate pool grows from 100K to 850K motions.}
  \label{fig:retrieval-efficiency}
\end{figure}

\noindent\textbf{Efficiency Comparison.}\
We expand the HumanML3D candidate pool to evaluate online retrieval at scale and select SGAR~\cite{zhang2025sgar} as a representative embedding-based retrieval method. As shown in Figure~\ref{fig:retrieval-efficiency}, SGAR's throughput drops from 40.19 to 5.67 queries per second (QPS) as the pool grows. This decline stems from exhaustive similarity scoring. SGAR compares each query with all candidate motion embeddings, so larger pools require more similarity computations. HiMoID decodes three-token semantic IDs with a fixed beam width using a prefix tree built offline, avoiding exhaustive candidate scoring. It maintains about 14.6 QPS across the tested sizes, reaching about \(2.6\times\) SGAR's throughput at 850K candidates. This scale invariance makes the generative paradigm more predictable and scalable in large-scale retrieval settings. The proposed components are used only during training, so our model shares the generative baseline's generator architecture and decoding procedure, yielding similar online throughput.

\subsection{Ablation Study}
\label{sec:ablation}

\noindent\textbf{Ablation of Key Components.}\
We conduct ablation experiments on HumanML3D and KIT-ML to evaluate the contribution of each component. We adapt prior generative retrieval approaches~\cite{rajput2023tiger,zhang2024irgen,kim2025genius} to text-to-motion retrieval to establish a common baseline. After tuning the baseline's hyperparameters, we use the same text and motion encoder architectures, residual quantizer configuration, text-conditioned autoregressive generator architecture, and training and inference settings across variants, except for changes required by the components under study. We evaluate motion reconstruction (MR) and semantic prefix organization (SPO) individually and in combination, then add tree-distribution supervision (TDS) to the configuration with both MR and SPO.

\begin{table}[!t]
  \centering
  \caption{Component ablation on HumanML3D and KIT-ML. MR, SPO, and TDS denote motion reconstruction, semantic prefix organization, and tree-distribution supervision. Checkmarks indicate enabled components.}
  \label{tab:component-ablation}
  \small
  \setlength{\tabcolsep}{2pt}
  \begin{tabular*}{\linewidth}{@{\extracolsep{\fill}}ccccccccc@{}}
    \toprule
    \multicolumn{3}{c}{Components} & \multicolumn{3}{c}{HumanML3D} & \multicolumn{3}{c}{KIT-ML} \\
    \cmidrule(lr){1-3}\cmidrule(lr){4-6}\cmidrule(lr){7-9}
    MR & SPO & TDS & R@1$\uparrow$ & R@5$\uparrow$ & R@10$\uparrow$ & R@1$\uparrow$ & R@5$\uparrow$ & R@10$\uparrow$ \\
    \midrule
    & & & 10.39 & 27.40 & 39.00 & 15.18 & 34.82 & 51.08 \\
    $\checkmark$ & & & 11.21 & 30.50 & 42.79 & 16.75 & 36.14 & 54.58 \\
    & $\checkmark$ & & 11.96 & 31.58 & 42.63 & 17.11 & 36.87 & 55.30 \\
    $\checkmark$ & $\checkmark$ & & 13.97 & 34.59 & 47.56 & 18.80 & 40.48 & 57.60 \\
    $\checkmark$ & $\checkmark$ & $\checkmark$ & \textbf{14.73} & \textbf{37.65} & \textbf{51.10} & \textbf{19.52} & \textbf{44.94} & \textbf{60.12} \\
    \bottomrule
  \end{tabular*}
\end{table}

\begin{figure}[!t]
  \centering
  \includegraphics[width=\linewidth]{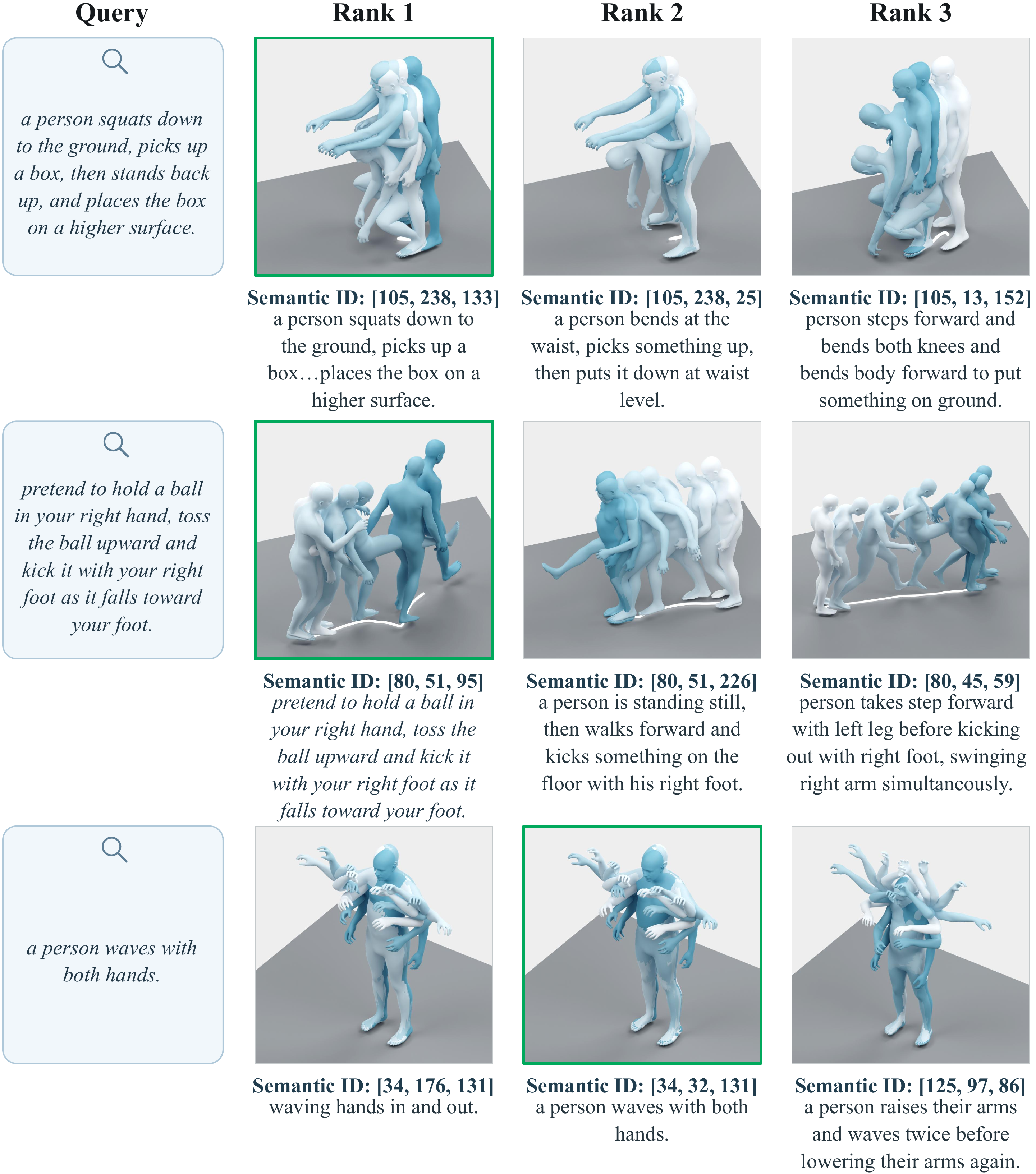}
  \caption{Top-three text-to-motion retrieval results with motion descriptions and semantic IDs. Green borders mark correct matches.}
  \label{fig:qualitative-retrieval}
\end{figure}

As shown in Table~\ref{tab:component-ablation}, adding either MR or SPO individually improves retrieval performance, confirming the effectiveness of motion reconstruction and semantic prefix constraints within our framework. MR supervises the quantized representation through full-motion reconstruction, incorporating motion content directly into semantic ID learning. SPO constrains prefix sharing among semantically related motions, incorporating their semantic relationships into candidate organization. Combining the two components outperforms using either one alone, further confirming the value of constraining both motion content and candidate relationships.

With TDS, the full model achieves the best results on all three metrics, confirming that branch distribution supervision further improves generative retrieval performance. This gain stems from TDS explicitly supervising the probability mass of relevant branches alongside the autoregressive objective, providing training signals for candidates beyond the annotated path.

\FloatBarrier

\subsection{In-depth Analysis}
\label{sec:visualization}

\noindent\textbf{Qualitative Retrieval Results.}\
Figure~\ref{fig:qualitative-retrieval} shows three text queries and their top three retrieved motions, annotated with descriptions and semantic IDs. The correct motions rank first for the object pickup-and-placement and kicking queries, and second for the waving query, whose other candidates also contain related waving movements. These results illustrate the model's ability to match query semantics. The retrieved candidates also reveal a connection between shared prefixes and motion semantics. For example, the first two motions in the first row both involve picking up and placing an object and share the prefix [105, 238], but use squatting and bending postures, respectively, and have different final tokens. Similarly, the first two motions in the second row share the prefix [80, 51] while retaining the distinction between kicking a falling ball and kicking an object on the floor. These examples illustrate how semantic IDs organize related motions through shared prefixes while retaining specific motion differences in subsequent tokens.

\begin{figure}[!t]
  \centering
  \includegraphics[width=\linewidth]{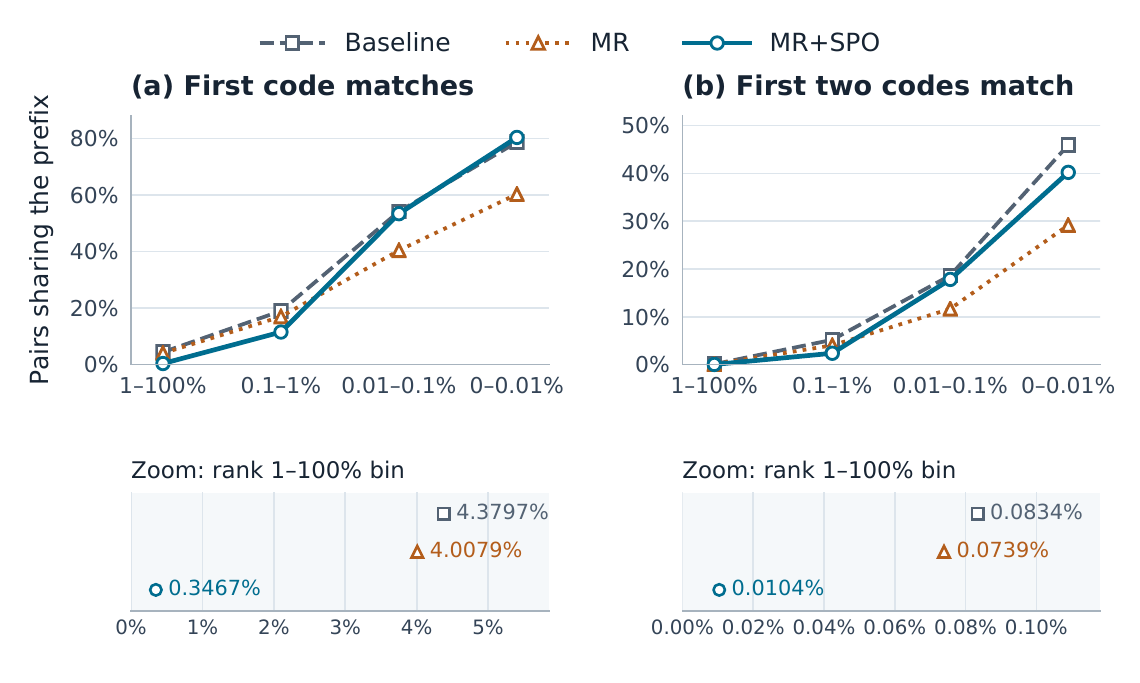}
  \caption{Prefix sharing by caption similarity on HumanML3D. Each bin shows the percentage of pairs sharing (a) the first code or (b) the first two codes. Similarity increases left to right. Lower panels magnify the least similar 99\% of pairs.}
  \label{fig:prefix-sharing}
  \par\medskip
  \begin{minipage}{\linewidth}
    \centering
    \captionof{table}{Retrieval recall (\%) for different per-level codebook sizes \(K\).}
    \label{tab:codebook-size}
    \small
    \setlength{\tabcolsep}{3pt}
    \begin{tabular*}{\columnwidth}{@{\extracolsep{\fill}}crrrrrr@{}}
      \toprule
      & \multicolumn{3}{c}{HumanML3D} & \multicolumn{3}{c}{KIT-ML} \\
      \cmidrule(lr){2-4}\cmidrule(lr){5-7}
      \(K\) & R@1 & R@5 & R@10 & R@1 & R@5 & R@10 \\
      \midrule
      32 & 9.18 & 26.44 & 37.49 & 16.39 & 38.43 & 54.58 \\
      64 & 12.01 & 33.90 & 44.75 & 17.35 & 41.57 & 57.59 \\
      128 & 14.38 & 36.89 & 50.11 & 18.43 & 43.01 & 59.52 \\
      256 & 14.73 & \textbf{37.65} & \textbf{51.10} & \textbf{19.52} & 44.94 & \textbf{60.12} \\
      512 & \textbf{14.77} & 37.58 & 50.37 & 18.80 & 43.49 & 58.07 \\
      1024 & 14.63 & 37.49 & 51.03 & 19.04 & \textbf{45.66} & 59.16 \\
      \bottomrule
    \end{tabular*}
  \end{minipage}
\end{figure}

\noindent\textbf{Semantic ID Prefix Sharing Analysis.}\
Figure~\ref{fig:prefix-sharing} extends the qualitative examples to pairs of distinct motions in the HumanML3D test gallery. We rank the pairs by the cosine similarity of their captions using a fixed text encoder independent of training. The comparison with the baseline demonstrates the effect of motion reconstruction (MR) on prefix assignments. Adding MR reduces prefix sharing at both depths across all four similarity bins. This decrease is consistent with the reconstruction objective, which encourages the quantized representation to distinguish motions that are semantically similar but performed differently. Relative to MR alone, adding SPO increases prefix sharing at both depths in the two highest similarity bins and decreases it in the other two bins. These results highlight the distinct roles of MR and SPO in learning semantic IDs. MR aims to preserve fine-grained motion features for precise retrieval, while SPO concentrates prefix sharing among motions with highly similar captions.

\noindent\textbf{Semantic ID Collision Analysis.}\  To assess whether collisions affect recall, we analyze the HumanML3D test gallery. Of all gallery motions, 8.6\% share an ID with another motion. Under exact caption matching, 69.5\% of collision buckets are benign, containing only positives for the same query. Randomizing within-bucket order has little effect on recall.

\noindent\textbf{Codebook Size Analysis.}\  A larger codebook offers more choices at each quantization level. Table~\ref{tab:codebook-size} shows consistent gains as the codebook grows to 256 entries per level, suggesting that retrieval benefits from the added capacity over this range. Beyond that point, gains appear only at individual cutoffs. We therefore use 256 codewords per level on both datasets.

\begin{table}[!tb]
  \centering
  \caption{Retrieval recall (\%) for different quantization depths \(L\).}
  \label{tab:quantization-depth}
  \small
  \setlength{\tabcolsep}{3pt}
  \begin{tabular*}{\columnwidth}{@{\extracolsep{\fill}}crrrrrr@{}}
    \toprule
    & \multicolumn{3}{c}{HumanML3D} & \multicolumn{3}{c}{KIT-ML} \\
    \cmidrule(lr){2-4}\cmidrule(lr){5-7}
    \(L\) & R@1 & R@5 & R@10 & R@1 & R@5 & R@10 \\
    \midrule
    2 & 12.08 & 31.14 & 43.52 & 16.51 & 39.52 & 55.42 \\
    3 & \textbf{14.73} & \textbf{37.65} & \textbf{51.10} & \textbf{19.52} & \textbf{44.94} & \textbf{60.12} \\
    4 & 14.11 & 36.53 & 50.05 & 18.80 & 43.13 & 58.31 \\
    5 & 13.97 & 36.21 & 48.74 & 18.19 & 42.77 & 58.67 \\
    6 & 13.61 & 36.46 & 48.52 & 17.11 & 41.69 & 57.35 \\
    \bottomrule
  \end{tabular*}
\end{table}

\noindent\textbf{Quantization Depth Analysis.}\  Quantization depth controls both the granularity of motion IDs and the number of autoregressive decoding steps. Table~\ref{tab:quantization-depth} shows that adding a third level improves recall on both datasets. Beyond three levels, the extra tokens require more decoding steps, and recall stays below that of the three-level setting at every cutoff. We therefore adopt three levels on both datasets.

\begin{table}[!tb]
  \centering
  \caption{Retrieval recall (\%) for different beam sizes.}
  \label{tab:beam-size}
  \small
  \setlength{\tabcolsep}{3pt}
  \begin{tabular*}{\columnwidth}{@{\extracolsep{\fill}}crrrrrr@{}}
    \toprule
    & \multicolumn{3}{c}{HumanML3D} & \multicolumn{3}{c}{KIT-ML} \\
    \cmidrule(lr){2-4}\cmidrule(lr){5-7}
    Beam & R@1 & R@5 & R@10 & R@1 & R@5 & R@10 \\
    \midrule
    10 & 13.45 & 34.95 & 47.90 & 16.63 & 40.96 & 53.37 \\
    20 & \textbf{14.73} & \textbf{37.65} & 51.10 & 19.52 & 44.94 & \textbf{60.12} \\
    30 & 14.29 & 36.94 & 51.37 & 19.28 & 43.73 & 59.64 \\
    40 & 13.84 & 36.64 & 51.12 & 19.76 & 44.10 & 57.95 \\
    50 & 14.57 & 35.48 & 51.58 & 19.76 & \textbf{45.30} & 58.80 \\
    60 & 14.59 & 36.05 & \textbf{51.80} & \textbf{20.00} & 44.34 & 58.67 \\
    \bottomrule
  \end{tabular*}
\end{table}

\noindent\textbf{Beam Size Analysis.}\  A wider beam keeps more candidate prefixes available during decoding. Table~\ref{tab:beam-size} shows that widening the beam from 10 to 20 consistently improves recall on both datasets. Further widening yields gains at some cutoffs and can increase the number of prefix extensions to score at each step. We use a beam size of 20 to balance recall and decoding cost.

%% file: sec/5_conclusion.tex
\section{Conclusion}
\label{sec:conclusion}

In this paper, we presented HiMoID, a generative framework for text-to-motion retrieval that formulates retrieval as query-conditioned generation of hierarchical semantic IDs. We introduced hierarchical semantic motion quantization and retrieval-oriented semantic ID generation to enable precise text-to-motion retrieval. Extensive experiments on HumanML3D and KIT-ML demonstrate that HiMoID achieves strong performance in text-to-motion retrieval.